\documentclass[10pt,twocolumn,letterpaper]{article}

\usepackage[pagenumbers]{cvpr} 

\usepackage{capt-of}
\usepackage[dvipsnames]{xcolor}
\usepackage{xspace}
\usepackage{enumitem}
\usepackage{amsmath,amssymb,amsbsy,amsfonts,dsfont,pifont,bm,bbm,mathrsfs,mathtools,nicefrac}
\usepackage{algorithm,algpseudocode,listings}
\usepackage{booktabs,multirow,adjustbox,diagbox,threeparttable,tabularray}

\newcommand{\tocite}[1]{{\color{red} [TO CITE]}}
\newcommand{\method}{UniReal\xspace}

\definecolor{cvprblue}{rgb}{0.21,0.49,0.74}
\usepackage[pagebackref,breaklinks,colorlinks,citecolor=cvprblue]{hyperref}
\usepackage{wrapfig}
\usepackage[capitalize]{cleveref}  
\crefname{section}{Sec.}{Secs.}
\Crefname{section}{Section}{Sections}
\crefname{table}{Tab.}{Tabs.}
\Crefname{table}{Table}{Tables}
\crefname{figure}{Fig.}{Figs.}
\Crefname{figure}{Figure}{Figures}
\crefname{equation}{Eq.}{Eqs.}
\Crefname{equation}{Equation}{Equations}
\def\paperID{3816}
\def\confName{CVPR}
\def\confYear{2025}

\title{UniReal: \underline{Uni}versal Image Generation and Editing  \\ via Learning \underline{Real}-world Dynamics }

\author{
    Xi Chen$^{1}$ \quad
    Zhifei Zhang$^{2}$ \quad
    He Zhang$^{2}$ \quad
    Yuqian Zhou$^{2}$ \quad
    Soo Ye Kim$^{2}$ \quad
    Qing Liu$^{2}$ \quad
    Yijun Li$^{2}$ \quad 
    \\
    Jianming Zhang$^{2}$ \quad
    Nanxuan Zhao$^{2}$ \quad
    Yilin Wang$^{2}$ \quad
    Hui Ding$^{2}$ \quad
    Zhe Lin$^{2}$ \quad
    Hengshuang Zhao$^{1}$\\[5pt]
    $^{1}$The University of Hong Kong \quad
    $^{2}$Adobe \quad
}

\begin{document}
\twocolumn[{
\renewcommand\twocolumn[1][]{#1}
\maketitle
\begin{center}
    \vspace{-16pt}
    \includegraphics[width=1.0\linewidth]{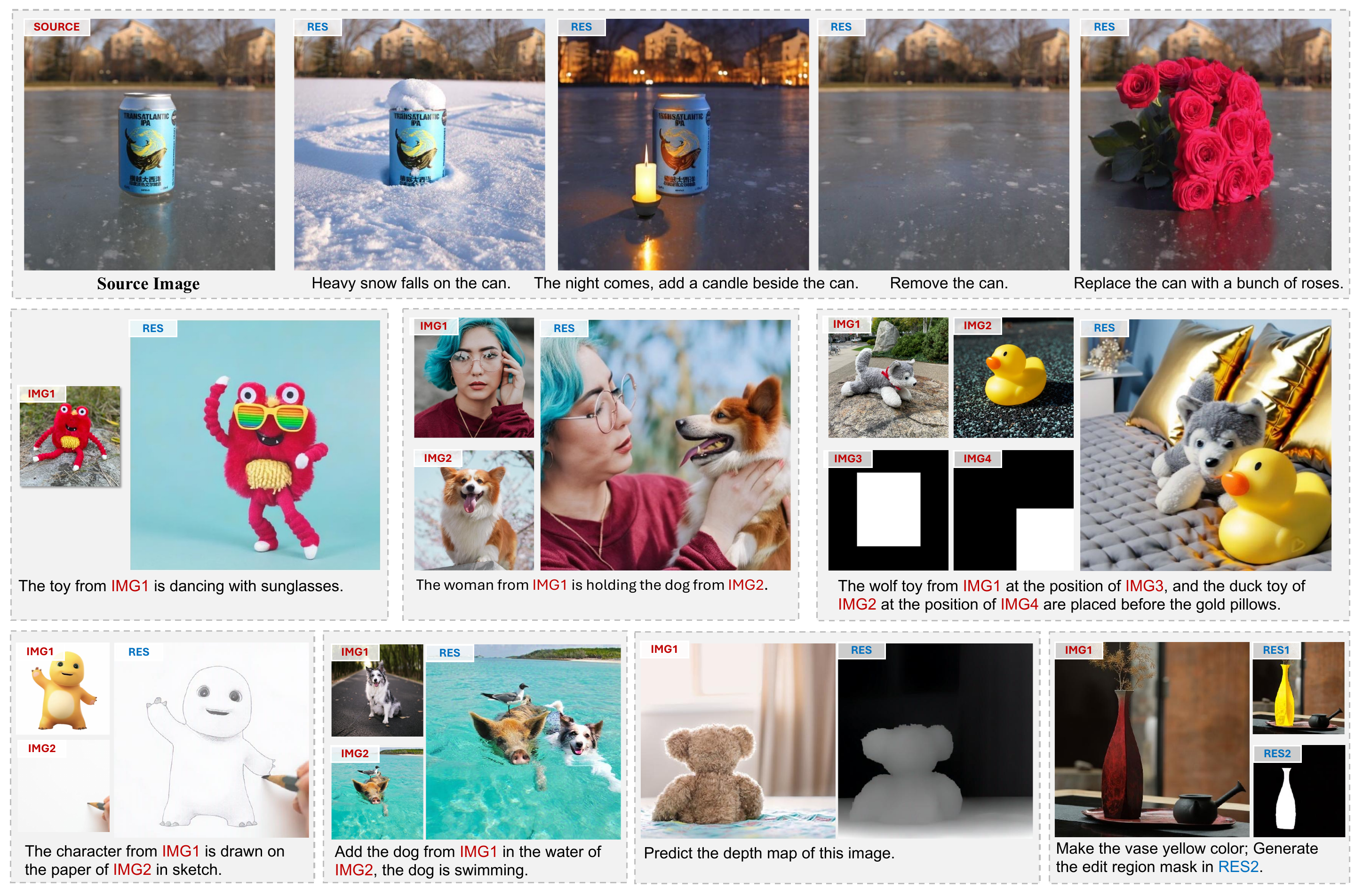}
    \vspace{-20pt}
    \captionsetup{type=figure}
    \caption{%
    \textbf{Demonstrations of UniReal’s versatile capabilities.}
    As a \textit{universal} framework, \method supports a broad spectrum of image generation and editing tasks within a single model, accommodating diverse input-output configurations and generating highly \textit{realistic} results, which effectively handle challenging scenarios, \eg, shadows, reflections, lighting effects, object pose changes, \etc.
    }
    \label{fig:teaser}
    \vspace{-2pt}
\end{center}
}]

\begin{abstract}
\vspace{-8pt}
We introduce \textbf{\method}, a unified framework designed to address various image generation and editing tasks.
Existing solutions often vary by tasks, yet share fundamental principles: preserving consistency between inputs and outputs while capturing visual variations. 
Inspired by recent video generation models that effectively balance consistency and variation across frames, we propose a unifying approach that treats image-level tasks as discontinuous video generation.
Specifically, we treat varying numbers of input and output images as frames, enabling seamless support for tasks such as image generation, editing, customization, composition, etc.
Although designed for image-level tasks, we leverage videos as a scalable source for universal supervision. 
\method learns world dynamics from large-scale videos, demonstrating advanced capability in handling shadows, reflections, pose variation, and object interaction, while also exhibiting emergent capability for novel applications.  
The page of this project is \href{https://xavierchen34.github.io/UniReal-Page/}{here}.
\vspace{-15pt}
\end{abstract}

\section{Introduction}\label{sec:intro}
\vspace{-5pt}
The field of visual content creation has advanced significantly with the development of diffusion models~\cite{ho2020ddpm,rombach2022stablediffusion}, enabling a broad range of applications in image generation~\cite{podell2023sdxl,zhang2023controlnet,chen2023pixart,huang2023composer,mou2024t2i} and editing~\cite{brooks2023instructpix2pix,zhang2024magicbrush,chen2024anydoor,ye2023ip}.
However, as practical demands increase, the applications are becoming progressively specialized in both tasks and methods. This limits the ability to learn generalizable knowledge across fields and increases the workload for designing task-specific methods and collecting domain-specific data. 
In this work, we explore designing a universal framework that could unify diverse tasks into a generalized formulation along with a scalable training paradigm.

Across diverse image generation and editing tasks, we observe shared core requirements, such as preserving \textit{consistency} between input and output images while bringing in controlled visual \textit{variation}. These requirements guide our design of a universal framework. Notably, video generation models like Sora type methods ~\cite{Sora,polyak2024moviegen,hong2022cogvideo,opensora,yang2024cogvideox} effectively balance frame \textit{consistency} with motion \textit{variation}, aligning closely with our goals. Thus, we adapt their design principles for image-level tasks, reformulating various image generation and editing tasks into a unified framework for    ``discontinuous'' frame generation.
Specifically, we introduce \textbf{\method}, a universal solution tackling different tasks within one diffusion transformer.
\method builds on the foundational structure of video generation models, using full attention to model relationships across frames.

UniReal treats varying numbers of input and output images as pseudo-frames, guided by a text prompt to support a wide range of applications. 
To unify input image types in multiple tasks, \eg, text-to-image generation, controllable generation, instructive editing, and multi-subject customization, we distinguish input images as three pivots: 1) the target image to edit on, 2) reference image that carries objects or visual elements to insert or preserve, and 3) condition map for layout or shape regularization.
To coordinate multiple input images under a unified text prompt, we introduce image index embeddings that link each image to its corresponding prompt term. For task synergy, we design a hierarchical prompting scheme, layering context-level and image-level guidance onto the base prompt. Together, these techniques enable \method to seamlessly integrate and compose diverse tasks.

Instead of curating task-specific data, we explore seeking universal supervision.
We find that many types of variations~(\textit{e.g.,} add, remove, attribute changes, structural changes) are naturally covered between two discontinuous video frames. 
In this way, we leverage large-scale frame pairs with captions as instructive editing data, and build an automatic pipeline to construct data from videos for image customization and composition tasks.

As shown in \cref{fig:teaser},
\method exhibits superior capabilities for maintaining consistency and preserving the details with the input images.
Furthermore, it demonstrates strong potential for modeling natural variations and simulating world dynamics such as lighting, reflections, and object interactions.
In addition, after learning versatile tasks, \method shows some emerged abilities for novel applications without training data.

\section{Related Work}\label{sec:related}
\noindent \textbf{Diffusion-based image editing.} Image editing is a general and practical topic that covers 
a lot of applications. Initial works~\cite{mokady2023nulltextinversion, meng2021sdedit, kawar2023imagic, cao2023masactrl, hertz2022prompt2prompt} explore training-free or tuning-based solutions to edit the attributes of the source image with prompts. 
Later, InstructPix2Pix~\cite{brooks2023instructpix2pix} and other works~\cite{zhao2024ultraedit,hui2024hqedit,zhang2024magicbrush,yang2024editworld} train instructive image editing on constructed datasets. 
Meanwhile, another branch of work explores mask-based editing, \textit{e.g.,} regenerate the regions covered by the given mask~\cite{zhuang2023powerpaint, ju2024brushnet} or compose the region from one image to another~\cite{chen2024anydoor,yang2023paintbyexample,song2023objectstitch,song2024imprint,chen2024mimicbrush}. 
Customized generation task~\cite{li2024blipdiffusion,ruiz2023dreambooth,liu2023cones,liu2023cones2,kumari2023customdiffusion} requires one or multiple reference images as input to generate new images with the same subject. 
Besides, there are also application-oriented tasks like virtual try-on~\cite{kim2024stableviton,chen2024wear,xu2024ootdiffusion}, face personalization~\cite{li2024photomaker,wang2024instantid,zhang2024flashface}, image stylization~\cite{sohn2023styledrop,wang2024instantstyle}, \textit{etc}. 
In general, we observe that the settings and solutions for image editing tasks vary significantly.  
However, there are still common requirements like keeping the consistency with the input images and modeling the visual variations according to specific conditions. 
In this work, we explore designing a universal framework and constructing generalizable supervision supporting versatile tasks.

\noindent \textbf{Universal generative model.} The omni solution for different image generation/editing tasks is a challenging topic. SEED-X~\cite{ge2024seedx} and Emu2~\cite{sun2024emu2} use an autoregressive model to predict the next tokens and use a separated diffusion model to generate the images. 
Tranfusion~\cite{zhou2024transfusion} predicts discrete tokens to for text and uses continuous vectors to represent images.
Show-o~\cite{xie2024show} leverages causal attention to process text tokens and uses full attention for image tokens.
VILA-U~\cite{wu2024vila} adopts discrete tokens for both text and image and leverages different decoders for different modalities.  
Nevertheless, as they mainly focus on understanding tasks, their image generation/editing abilities are not quite satisfactory as a side product. 
Some works explore a unified solution focusing on image generation and editing.
Instruct-Imagen~\cite{hu2024instructimagen} unifies image generation tasks together using multi-modal instructions.  
Pixwizard~\cite{lin2024pixwizard} proposes task embeddings for image editing and understanding tasks.
OmniGen~\cite{xiao2024omnigen} tokenizes the texts and images as a long tensor, using causal attention for text tokens and bidirectional attention for image tokens.  
ACE~\cite{han2024ace} designs the conditioning unit to receive different kinds of input images and uses a transformer to deal with multiple inputs.

\begin{figure*}[t]
\centering 
\includegraphics[width=1\linewidth]{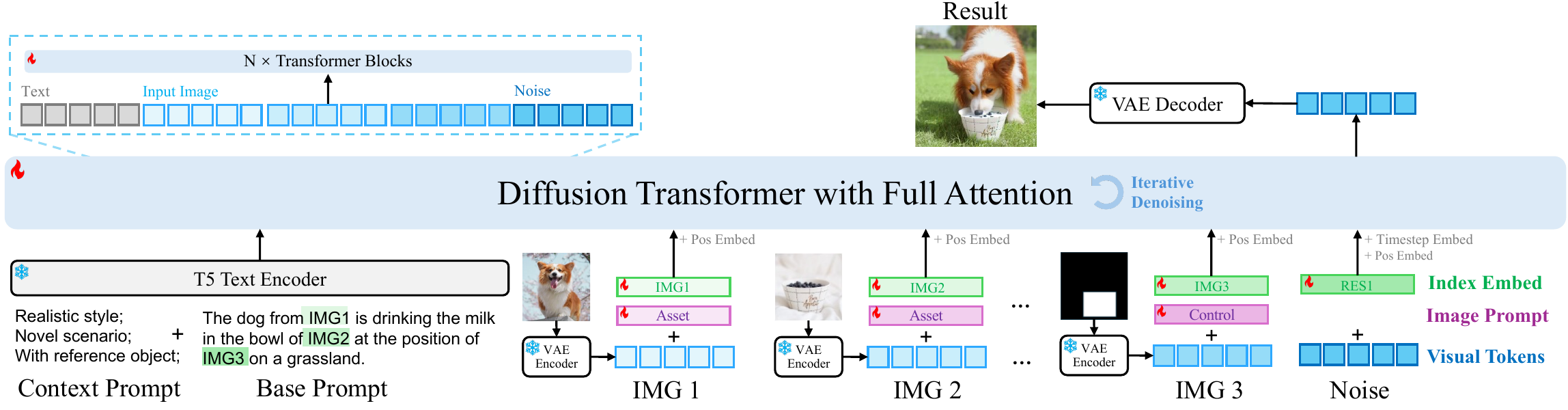} 
\vspace{-15pt}
\caption{
    \textbf{Overall pipeline of \method.}
    We formulate image generation and editing tasks as discontinuous frame generation.
    First, input images are encoded into latent space by VAE encoder. 
    Then, we patchify the image latent and noise latent into visual tokens. 
    Afterward, we add index embeddings and \textit{image prompt~(asset/canvas/control)} to the visual tokens.   
    At the same time, the \textit{context prompt} and \textit{base prompt} are processed by the T5 encoder. 
    We concatenate all the latent patches and text embeddings as a long 1D tensor and send them to the transformer. 
    Finally, we decode the denoised results to get the desired output images. 
}
\label{fig:pipeline}
\vspace{-10pt}
\end{figure*}

\section{Method}\label{sec:method}

\method leverages a video generation framework that processes arbitrary numbers of input and output images as video frames, naturally unifying multiple image generation and editing tasks. Furthermore, \method seeks universal supervision for these tasks from scalable video data, achieving highly realistic generation results. We will detail architecture design (\cref{sec:model}), data construction (\cref{sec:data}), and training schemes (\cref{sec:training}).

\subsection{Model Design}\label{sec:model}
This section will dive into the key designs of \method, \ie, text-image association and hierarchical prompt based on diffusion transformer.

\noindent \textbf{Diffusion transformer.} 
As illustrated in \cref{fig:pipeline}, 
\method treats input/output images as video frames and uses prompts to manage different tasks. 
Specifically, images are encoded to latent space by the VAE encoder, then those latent maps are 
Upon visual tokens, we add index embeddings to distinguish the image order and add the image prompt to indicate whether an image serves as, \eg, canvas/background or asset/foreground object. \cref{fig:ab_hie_prompt} illustrates the effect of image prompt. 
The position embeddings are added to each image/noise token, and the timestep embeddings are added to the noise tokens.  
At the same time, the text prompts are sent to a T5~\cite{T5} encoder to extract text tokens. 
We concatenate the image and noise tokens along with the text tokens as a long 1D tensor and send it to a transformer. 
The transformer uses full attention to model the relationship between images and the text prompt.

\noindent \textbf{Text-image association.} 
To refer to specific images in the text prompt,
we construct a set of embedding pairs to associate the visual tokens with corresponding texts.
Specifically, we use referring words like ``IMG1'' and ``IMG2'' to refer to the input images, and ``RES1'' and ``RES2'' for the output images.
We add them as special tokens for the T5 tokenizer. 
At the same time,  we learn image index embeddings for each referring word, which are added to the tokens of the corresponding image. 

\noindent \textbf{Hierarchical prompt.} 
Different tasks/datasets handle the same input differently. For example, image editing keeps the layout of the input image and makes local changes. However, with the same prompt, image customization generates a novel scenario and only preserves the reference object. This introduces the ambiguity for both the training and inference phases.      
To reduce the ambiguity when mixing multiple tasks and data sources, we propose hierarchical prompt. 
Besides the base prompts like ``put this dog on a grassland'', we design additional context prompts and image prompts to provide detailed indications. 

Context prompt provides attribute tags for different tasks and data sources, like
``realistic/synthetic data'', ``static/dynamic senario'', ``with reference object'', etc. 
Different from task embeddings that are used in previous works~\cite{sheynin2024emuedit,lin2024pixwizard}, some of our ``keywords'' can be shared among tasks, forcing those tasks to learn common features. Besides, as texts are naturally composable, we can easily compose different context prompts to realize novel functions.

The image prompt indicates the specific role of input images.
We split input images into three categories: \textit{canvas image, asset image, and control image}.
The canvas image serves as a background for the editing target with a fixed layout.
The asset image provides the reference objects or visual elements for image customization or composition, for which the model should implicitly conduct segmentation, and simulate the size/position/pose changes for the object.
The control image includes the mask/edge/depth map that regularizes the layout or shape.
The model should take distinctive actions for images of different categories. Therefore, we design learnable category embeddings and add them to the corresponding image tokens as ``image prompt''. 

During inference, the context prompt and image prompt can be automatically analyzed from the base prompt with a default value.
Thus, \method does not require extra effort on writing prompts. Nevertheless, users can manually revise the task and image prompt to get more novel effects.

\subsection{Dataset Construction}\label{sec:data}
Some previous works~\cite{yang2024editworld,zhao2024ultraedit,hui2024hqedit,OMNIEDIT} leverage complicated workflows to collect data for specific tasks. 
Differently, we start with video data and explore leveraging the natural consistency and variation between video frames to benefit various image generation and editing tasks.

\begin{figure}[t]
\centering 
\includegraphics[width=0.99\linewidth]{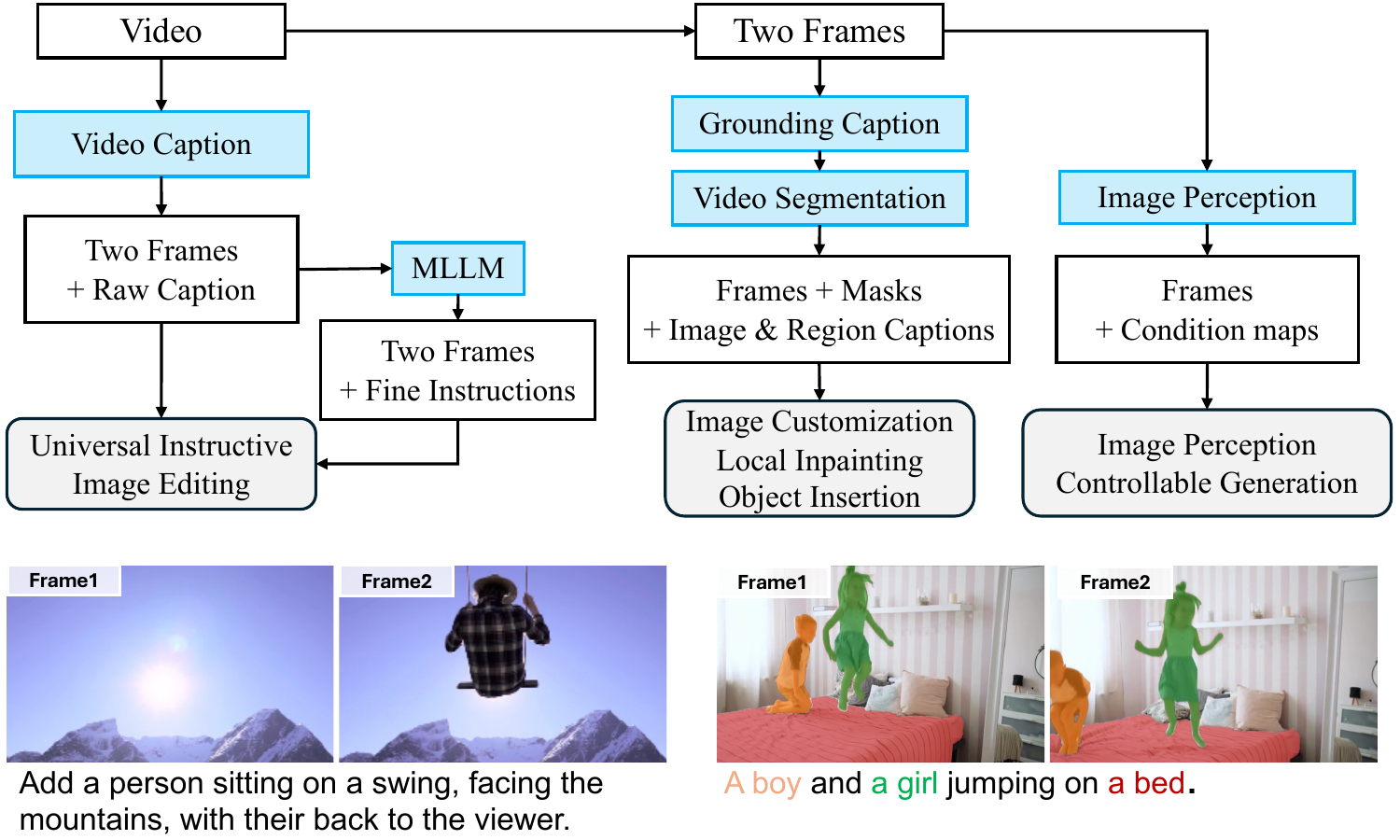} 
\vspace{-5pt}
\caption{%
    \textbf{Data construction pipeline.} 
    Starting from raw videos, we use off-the-shelf models to construct data for different kinds of tasks.
    Two examples of instructive editing and image customization data~(we segment objects from one frame to generate another frame) are given at the bottom of the image.
}
\label{fig:pipeline_data}
\vspace{-15pt}
\end{figure}

\noindent \textbf{Data construction pipline.} \cref{fig:pipeline_data} illustrates the data construction pipeline from raw videos to support various tasks. 
We first use a caption model to get the video-level captions. Then, we randomly pick two frames as before-/after-editing images and use video-level captions 
as instruction.  
We term this kind of data \textit{Video Frame2Frame}, and we observe that this data alone is already able to train a model with basic editing abilities. 
We further use GPT-4o mini~\cite{GPT-4o-mini} to get more precise instructions between two frames for a subset with 200K high-quality samples.

Besides, we use the grounding caption model, Kosmos-2~\cite{peng2023kosmos2} to generate image captions with the bounding boxes for the corresponding entities. 
Then, we use the boxes from one frame as prompts for SAM2~\cite{ravi2024sam2} to get the mask tracklet for the two frames.  In this way, we get the captions and masks for multiple objects in two video frames. This kind of data can support image customization~(\textit{Video Multi-object}), object insertion~(\textit{Video Object Insertion}), local inpainting~(\textit{Video ObjectAdd}), \textit{ect}. For example, for image customization, we segment each object from one frame and use them as reference images to generate another frame according to the caption. 
In addition, we reuse the masks and captions labeled by Kosmos-2 to support referring segmentation~(\textit{Video SEG}).
We also leverage image perception models~\cite{yang2024depthanythingv2,canny1986computational} to extract the depth map and the edge map, thus supporting controllable image generation and image perception~(\textit{Video Control}).

\begin{table}[t]
\caption{%
    \textbf{Statistics of datasets} used for training.
    We mix the existing datasets~(the first block) with our newly constructed 
    video-based datasets~(the second block). 
}
\label{tab:datasets}
\vspace{-7pt}
\centering\scriptsize
\setlength{\tabcolsep}{2.0pt}
\begin{tabular}{llcc}
\toprule
\textbf{No.} & \textbf{Dataset} & \textbf{\# Samples} & \textbf{Supporting Tasks}  \\
\midrule
1 & InstructP2P~\cite{brooks2023instructpix2pix}   & 300K   & Universal Instructive Editing  \\ 
2 & UltraEdit~\cite{zhao2024ultraedit}   & 500K   & Universal Instructive Editing  \\ 
3 & VTON-HD~\cite{choi2021vitonhd}   & 10K & Virtual Try-on  \\ 
4 & RefCOCO Series~\cite{refcoco+,mao2016refcocog}   & 150K  & Referring Segmentation \\ 
5 & T2I Generation~(in house)             & 300M   & Text to Image Generation  \\
6 & Instruct Editing~(in house)   & 2M   & Universal Instructive Editing  \\ 
7 & Object Insertion~(in house)   & 100K & Object Insertion with Reference  \\ 
\midrule
8 & Video Frame2Frame               & 8M   & Universal Instructive Editing  \\
9 & Video Multi-object   & 5M & Multi-subject Customization \\ 
10 & Video Object Insertion   & 1M & Object Insertion with Reference  \\ 
11 & Video ObjectAdd   & 1M & Object Insertion with Prompt  \\ 
12 & Video SEG   & 5M & Referring Segmentation  \\ 
13 & Video Control   & 3M & Perception, Controllable Generation \\ 
\bottomrule
\end{tabular}
\vspace{-15pt}
\end{table}

\noindent \textbf{Training data overview.} The datasets used for training are listed in \cref{tab:datasets}. Besides our constructed video-based datasets, we also use open-source data~\cite{brooks2023instructpix2pix,zhao2024ultraedit,choi2021vitonhd,refcoco+,mao2016refcocog} for specific tasks and our in-house datasets for instructive image editing and reference-based object insertion. 
Considering the difficulties of constructing image editing data, the public datasets are limited. Differently, our video-based data are easier to scale up. 

As explained in \cref{sec:model}, it is crucial to incorporate these datasets with context prompts. 
For example, in some videos from \textit{Video Frame2Frame}, there exist global changes~(\textit{e.g.}, background movements and camera motion) that are not captured by the instruction, which is not favored in instructive editing. In these cases, we analyze the optical flow and pixel MSE between frames to label each sample with ``static/dynamic scenario''. 
Besides, the instructive editing datasets~\cite{brooks2023instructpix2pix,zhao2024ultraedit} are in synthetic style, we tag them as ``synthetic style'' and give the context prompt of ``realistic style'' for real-image datasets. 
In addition, we give context prompts like ``with reference objects'' for \textit{Video Object Insertion}, and give ``perception task'' when training the model to predict masks or depth maps, \textit{etc.}

\begin{figure*}[t]
\centering 
\includegraphics[width=0.99\linewidth]{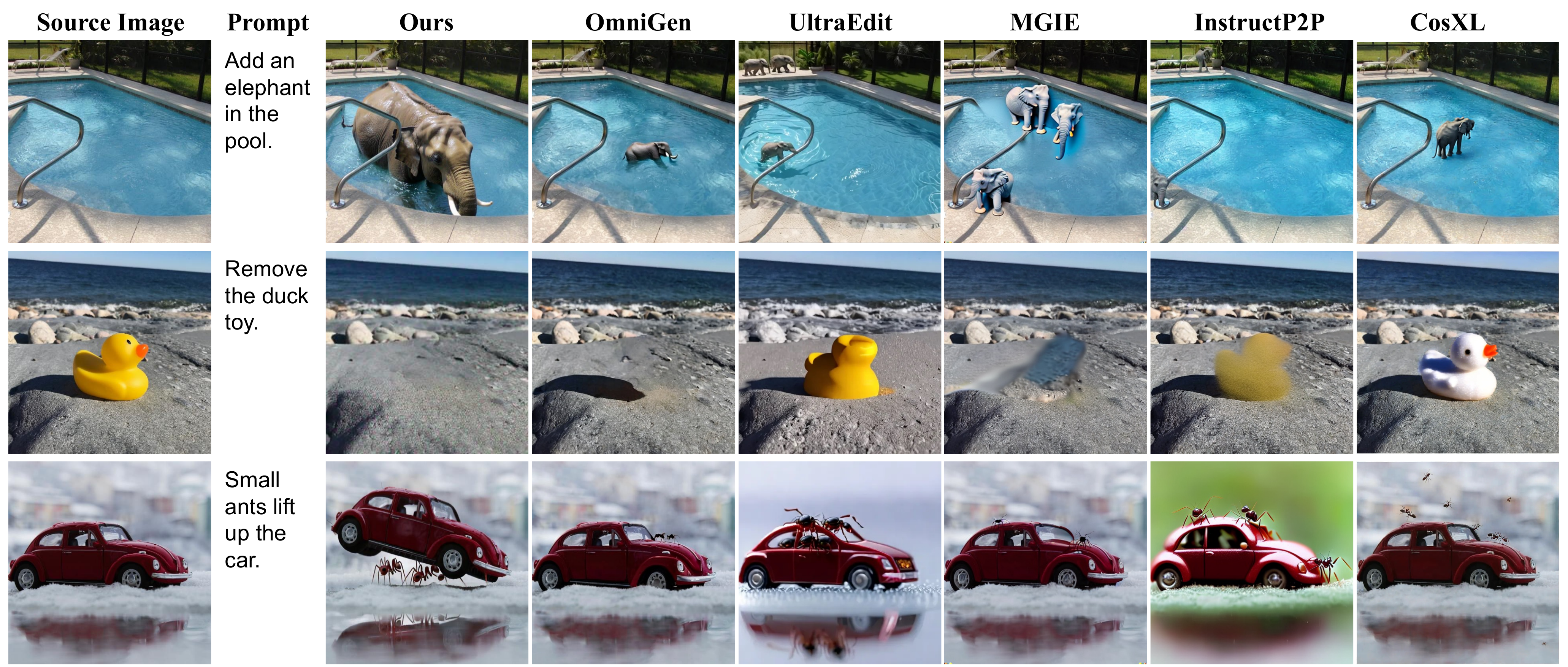} \vspace{-6pt}
\caption{%
    \textbf{Comparison results for instructive image editing.} 
    We compare with the state-of-the-art methods OmniGen~\cite{xiao2024omnigen}, UltraEdit~\cite{zhao2024ultraedit}, MGIE~\cite{fu2023mgie}, InstructPix2Pix~\cite{brooks2023instructpix2pix}, and CosXL~\cite{CosXL}.
    Our \method shows significant advantages in the aspects of instruction-following and generation quality. 
     We generate multiple results for each model and pick the best ones for demonstration.
    }
\label{fig:comp_edit}
\vspace{-10pt}
\end{figure*}

\vspace{-5pt}
\subsection{Training schemes} \label{sec:training}
\vspace{-5pt}
Our transformer model has 5B parameters. It is first pretrained with text-to-image and text-to-video data to get the basic generation ability under a small resolution of around 256$\times$256. 
Then, we train the model on all the datasets listed in \cref{tab:datasets} to learn multiple image generation/editing tasks~(256 resolution). 
Afterward, we progressively increase the resolution to 512 and 1024.
As we apply position embeddings for the image patches and the training images have different aspect ratios, \method could deal with different sizes and aspect ratios. 
We apply the learning rate of 1e-5 with a warm-up for each training stage. %
The training loss follows flow matching~\cite{lipman2022flowmatching}.

\section{Experiments}\label{sec:exp}
\begin{figure*}[t]
\centering 
\includegraphics[width=0.99\linewidth]{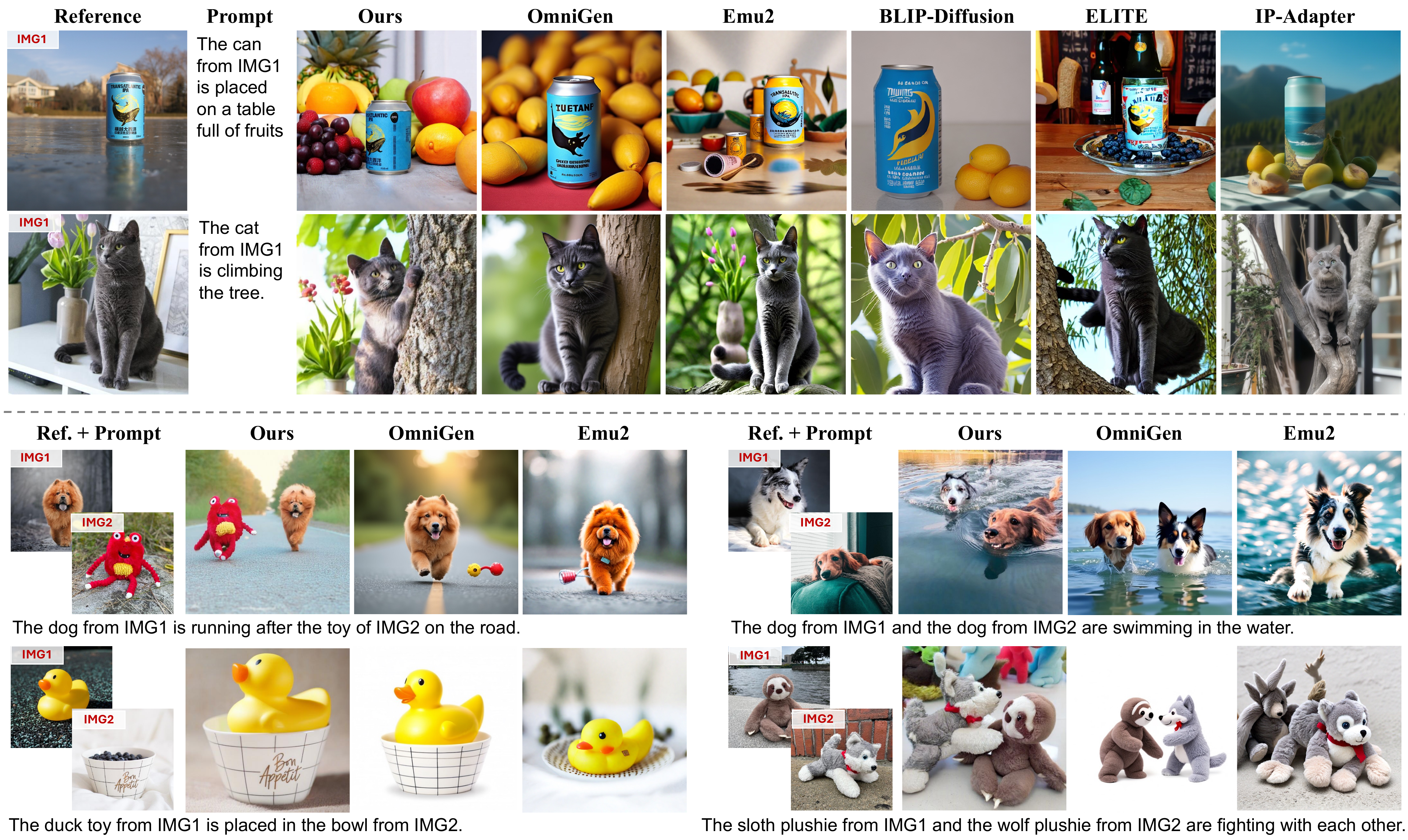} 
\vspace{-5pt}
\caption{%
    \textbf{Qualitative comparison for image customization.} 
    For single subject, we compare with OmniGen~\cite{xiao2024omnigen}, Emu2~\cite{sun2024emu2}, BLIP-Diffusion~\cite{li2024blipdiffusion}, ELITE~\cite{wei2023elite}, and IP-Adapter~\cite{ye2023ip} with Flux~\cite{Flux-IPadapter} backbone. 
    For multiple subjects, we chose OmniGen and Emu2 as competitors. 
    The listed prompts are in the formats of \method, and they are formulated according to the requirements of each method.
}
\label{fig:comp_cus}
\vspace{-10pt}
\end{figure*}

\subsection{Comparisons with Existing Works}
As a universal model, \method exhibits superior abilities for various image generation and editing tasks, even compared with existing task-specific models.     
We select three challenging tasks: instructive image editing, customized image generation, and object insertion as representatives to demonstrate and analyze the performance of \method. 

\setlength{\tabcolsep}{2pt}
{
\begin{table}[t]
\centering
\caption{\textbf{Comparison results for instructive image editing} on EMU Edit~\cite{sheynin2024emuedit} and MagicBrush~\cite{zhang2024magicbrush} test sets.
We list the task-specific models in the first block and some concurrent universal models in the second block. 
\vspace{-10pt}
}
\label{tab:eval_image_editing}
\scalebox{0.5}
{
\centering
\begin{tabular}{lccccc|ccccc}
    \toprule
     & \multicolumn{5}{c|}{ EMU Edit Test set} & \multicolumn{5}{c}{MagicBrush Test Set}\\
    \midrule
    Method & $\text{CLIP}_{dir}\!\uparrow$ &  $\text{CLIP}_{im}\!\uparrow$ & $\text{CLIP}_{out}\!\uparrow$ &  $\text{L1}\!\downarrow$  & DINO$\uparrow$  & $\text{CLIP}_{dir}\!\uparrow$ &  $\text{CLIP}_{im}\!\uparrow$ & $\text{CLIP}_{out}\!\uparrow$ &  $\text{L1}\!\downarrow$ & DINO$\uparrow$  \\
    \midrule
    InstructPix2Pix~\cite{brooks2023instructpix2pix} & 0.078 & 0.834 & 0.219 & 0.121 & 0.762 
                    & 0.115 & 0.837 & 0.245 & 0.093 & 0.767 \\
    MagicBrush~\cite{zhang2024magicbrush}     & 0.090 & 0.838 & 0.222 & 0.100 & 0.776 
                    & 0.123 & 0.883 & 0.261 & 0.058 & 0.871 \\
    PnP~\cite{tumanyan2023pnp}            & 0.028 & 0.521 & 0.089 & 0.304 & 0.153 
                    & 0.025 & 0.568 & 0.101 & 0.289 & 0.220 \\
    Null-Text Inv.~\cite{mokady2023nulltextinversion} & 0.101 & 0.761 & 0.236 & 0.075 & 0.678 
                    & 0.121 & 0.752 & 0.263  & 0.077 & 0.664 \\
    
    UltraEdit~\cite{zhao2024ultraedit}      & 0.107 & 0.793 & 0.283 & 0.071 & 0.844 
                    & -  & 0.868  & -  & 0.088  & 0.792 \\
                    
    EMU Edit~\cite{sheynin2024emuedit}       &0.109 & 0.859 & 0.231 & 0.094 & 0.819
                    &0.135 & 0.897 & 0.261 & \textbf{0.052} & \textbf{0.879} \\
    
    \midrule
    ACE~\cite{han2024ace}      & 0.086 & \textbf{0.895}  &  0.274  &  0.076 &  \textbf{0.862} 
                    & -  & -  & 0.284  & -  & - \\
    
    OmniGen~\cite{xiao2024omnigen}      & -  & 0.836 &  0.233 & -  & 0.804
                    & -  &- & -  & -  & - \\
    
    PixWizard~\cite{lin2024pixwizard}      & 0.104  & 0.845  & 0.248  & \textbf{0.069}  & 0.798 
                    & 0.124  & 0.884  & 0.265  & 0.063  & 0.876 \\

    \method~(ours)             & \textbf{0.127} & 0.851 & \textbf{0.285} & 0.099 & 0.790
                    & \textbf{0.151} & \textbf{0.903} & \textbf{0.308} & 0.081 & 0.837\\
    \bottomrule
\end{tabular}
}
\vspace{-10pt}
\end{table}
}

\noindent \textbf{Instructive image editing.} Users provide a prompt to edit the input image in free forms, like adding/removing an object, changing the attributes or styles, \textit{etc}. 
We demonstrate the qualitative comparisons in \cref{fig:comp_edit}, where we make comparisons with several state-of-the-art public models~\cite{xiao2024omnigen, zhao2024ultraedit,fu2023mgie, brooks2023instructpix2pix, CosXL}.
\method shows a clear advantage for dealing with challenging cases like simulating the size and status of an elephant in the water and removing the duck toy along with the shadows.
In the third row, \method successfully understands the interactions between the aunts and the car while modeling the car's reflections.

The quantitative results on EMU Edit~\cite{sheynin2024emuedit} and MagicBrush~\cite{zhang2024magicbrush} test sets are reported in \cref{tab:eval_image_editing}. 
$\text{CLIP}_{dir}$ evaluates the agreements between the changes of CLIP~\cite{CLIP} text and CLIP image embeddings.  $\text{CLIP}_{out}$ estimates the similarities between the editing results and the descriptions of the expected output. $\text{CLIP}_{im}$, $\text{DINO}$, and $\text{L1}$ denotes the CLIP similarity, DINO~\cite{DINO} similarity, and L1 distance between the editing results and the source images. 
\method achieves the best performance for $\text{CLIP}_{dir}$ and $\text{CLIP}_{out}$, and get the best $\text{CLIP}_{im}$ on Magicbrush test set.
As we expect obvious changes between the output and input, $\text{DINO}$ and $\text{L1}$ are not quite reasonable metrics. 
Considering our model accurately follows the instructions to make obvious edits, it is reasonable that the similarities between the editing results and the source image would be lower.

\begin{table}[t]
\caption{%
    \textbf{Quantitative results for customized generation} on DreamBench~\cite{ruiz2023dreambooth}. We report the oracle results in the first row and compare both tuning methods and zero-shot methods.
}
\label{tab:comp_cus}
\vspace{-7pt}
\centering\scriptsize
\setlength{\tabcolsep}{8.6pt}
\scalebox{1.00}
{
\begin{tabular}{lccc}
\toprule
\textbf{Model}   &  $\textbf{CLIP-T}\!\uparrow$   & $\textbf{CLIP-I}\!\uparrow$ &  $\textbf{DINO}\!\uparrow$   \\
\toprule
Oracle~(reference images) & - & 88.5 & 77.4 \\
\midrule
Textual Inversion~\cite{mokady2023nulltextinversion}   & 0.255 & 0.780 &  0.569  \\
DreamBooth~\cite{ruiz2023dreambooth} & 0.305 & 0.803 &  0.668  \\
BLIP-Diffusion~\cite{li2024blipdiffusion} & 0.302 & 0.805 &  0.670  \\
\midrule
ELITE~\cite{wei2023elite}  & 0.296  & 0.772 & 0.647  \\
Re-Imagen~\cite{chen2022reimagen} &  0.270  & 0.740 & 0.600  \\
BootPIG~\cite{purushwalkam2024bootpig} & 0.311 & 0.797 & 0.674  \\
SuTI~\cite{chen2024suti}  & 0.304 & \textbf{0.819} & \textbf{0.741} \\
OmniGen~\cite{xiao2024omnigen}~(our test)  & 0.320  & 0.810 & 0.693 \\ 
\method~(ours)   & \textbf{0.326}  & 0.806 & 0.702 \\
\bottomrule
\end{tabular}
}
\vspace{-15pt}
\end{table}

\noindent \textbf{Customized image generation.} 
Preserving the details of the reference object and following the novel text prompt simultaneously is a challenging task.  In \cref{fig:comp_cus}, we provide qualitative comparisons for customized image generation. 
We select examples from DreamBench~\cite{ruiz2023dreambooth} to conduct both single object customization~(top row) and multiple subject compositions~(bottom row). 
\method demonstrate significant advantages compared with other zero-shot models~\cite{xiao2024omnigen,sun2024emu2,li2024blipdiffusion,wei2023elite,ye2023ip} from all-round perspectives.  
Our model could precisely maintain the fine details of the logos on the can~(first row) and the berry bowl~(last row). 
Besides, \method could handle drastic changes like making the cat climb the tree or letting the dogs swim. 
Meanwhile, our method can also accurately preserve details while modeling interactions between different objects.

The quantitative results on DreamBench~\cite{ruiz2023dreambooth} are reported in \cref{tab:comp_cus}. CLIP-T measures the similarity between the generated images and the text prompts, CLIP-I and DINO measure image similarity between the generated and reference images. \method gets the best CLIP-T showing superior instruction-following ability. 
As some test prompts require to edit the object's attribute, it is a trade-off between the text following and the generation fidelity.
Even though, \method still gets competitive DINO and CLIP-I scores.

\begin{figure}[t]
\centering 
\includegraphics[width=0.99\linewidth]{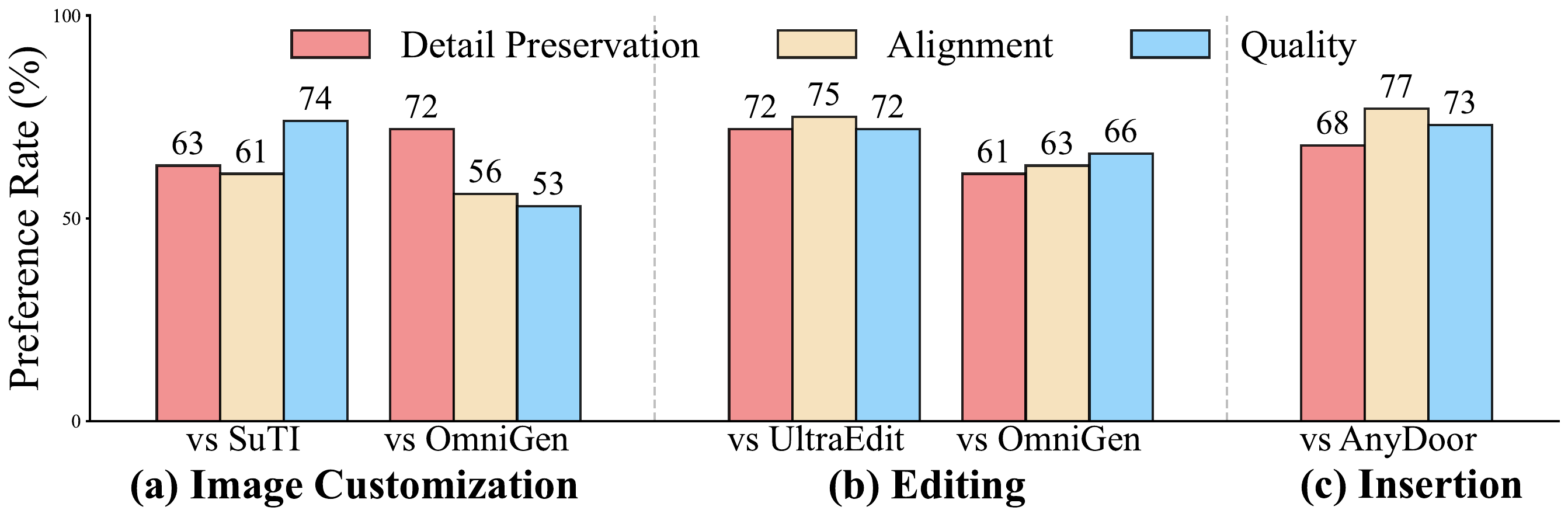} 
\vspace{-8pt}
\caption{%
    \textbf{Our preference rates against other methods} evaluated by user studies. 
    We compare SuTI~\cite{chen2024suti}, OmniGen~\cite{xiao2024omnigen}, UltraEdit~\cite{zhao2024ultraedit} and AnyDoor~\cite{chen2024anydoor} for different tasks.
}
\label{fig:user_study}
\vspace{-15pt}
\end{figure}

\begin{figure}[t]
\centering 
\includegraphics[width=0.99\linewidth]{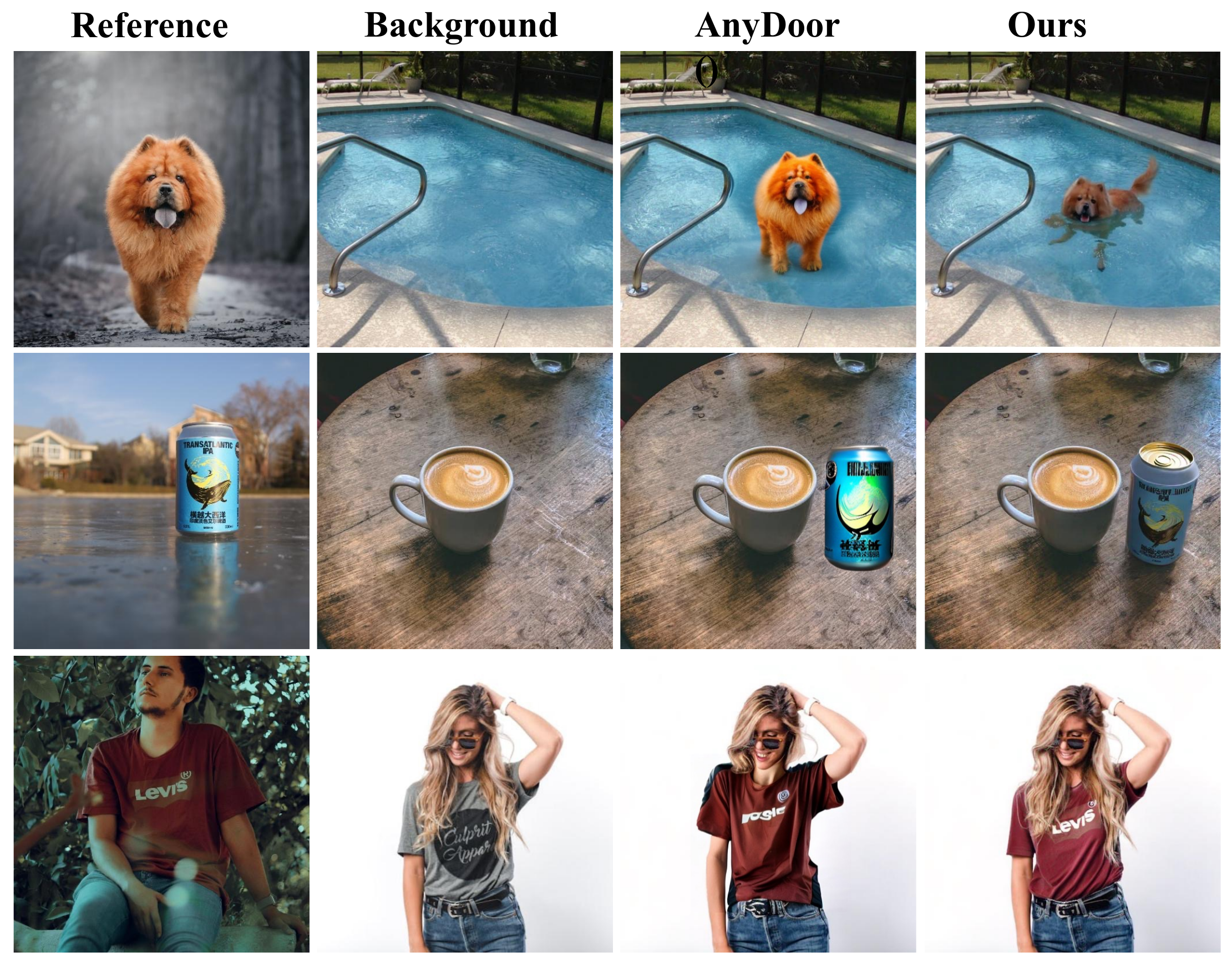} 
\vspace{-5pt}
\caption{%
    \textbf{Comparison results for object insertion.} 
    Our method could automatically adjust the status of the reference object according to the environment
    and strictly preserve the background.
    Our method does not require any mask as input.
}
\label{fig:comp_insert}
\vspace{-5pt}
\end{figure}

\begin{figure}[t]
\centering 
\includegraphics[width=0.99\linewidth]{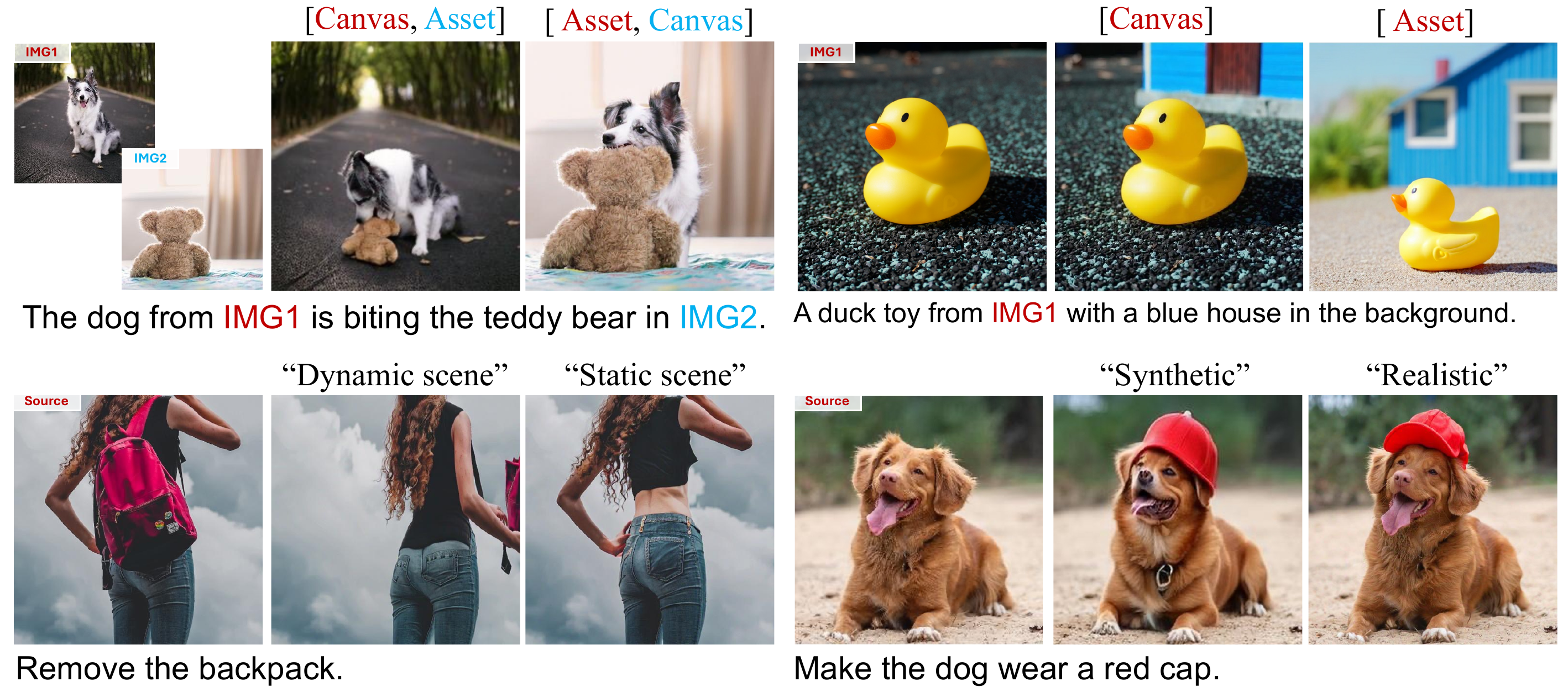} 
\vspace{-5pt}
\caption{%
    \textbf{Effects of hierarchical prompt.} 
    The same input could correspond to various types of targets when given different image prompts~(row~1) and context prompts~(row~2).
}
\label{fig:ab_hie_prompt}
\vspace{-10pt}
\end{figure}

\noindent \textbf{Reference-based object insertion.} In \cref{fig:comp_insert}, we compare \method with representative object insertion model, AnyDoor~\cite{chen2024anydoor}. 
AnyDoor requires additional masks for both the reference object and target location.
Differently, \method does not require any masks~(it could take additional masks to specify the target regions) but we give the corresponding text prompt like ``Add the dog from IMG1 to the swimming pool of IMG2.''
In this figure, we choose challenging cases that require intensive status changes, and 
observe that it is hard for AnyDoor~\cite{chen2024anydoor} to naturally put the dog into the water~(row~1) or automatically adjust the view of the can~(row~2).
Besides, as in the bottom row, a practical feature is that \method could strictly preserve the background pixels like the hair without any segmentation.

\noindent \textbf{User study.} 
In \cref{fig:user_study}, we report the human preference rates compared to each competitor. For image customization, we evaluate OmniGen~\cite{xiao2024omnigen} using 750 samples from DreamBench~\cite{ruiz2023dreambooth} and SuTI~\cite{chen2024suti} using 50 samples from their paper. For instruct editing and object insertion, we conduct comparisons with 100 challenging, self-collected samples.

We assign 10 annotators to evaluate three aspects: detail preservation, alignment, and image quality. Detail preservation assesses consistency with the reference object for image customization and object insertion, or with non-edited regions for instructive editing. Alignment measures how well the edit matches user intent, focusing on instruction-following for customization and editing, and background coherence for object insertion. Quality evaluates the accuracy and aesthetic appeal of the generated images. Overall, \method shows advantages over other methods in each aspect, with further details provided in the appendix.

\begin{figure}[t]
\centering 
\includegraphics[width=0.99\linewidth]{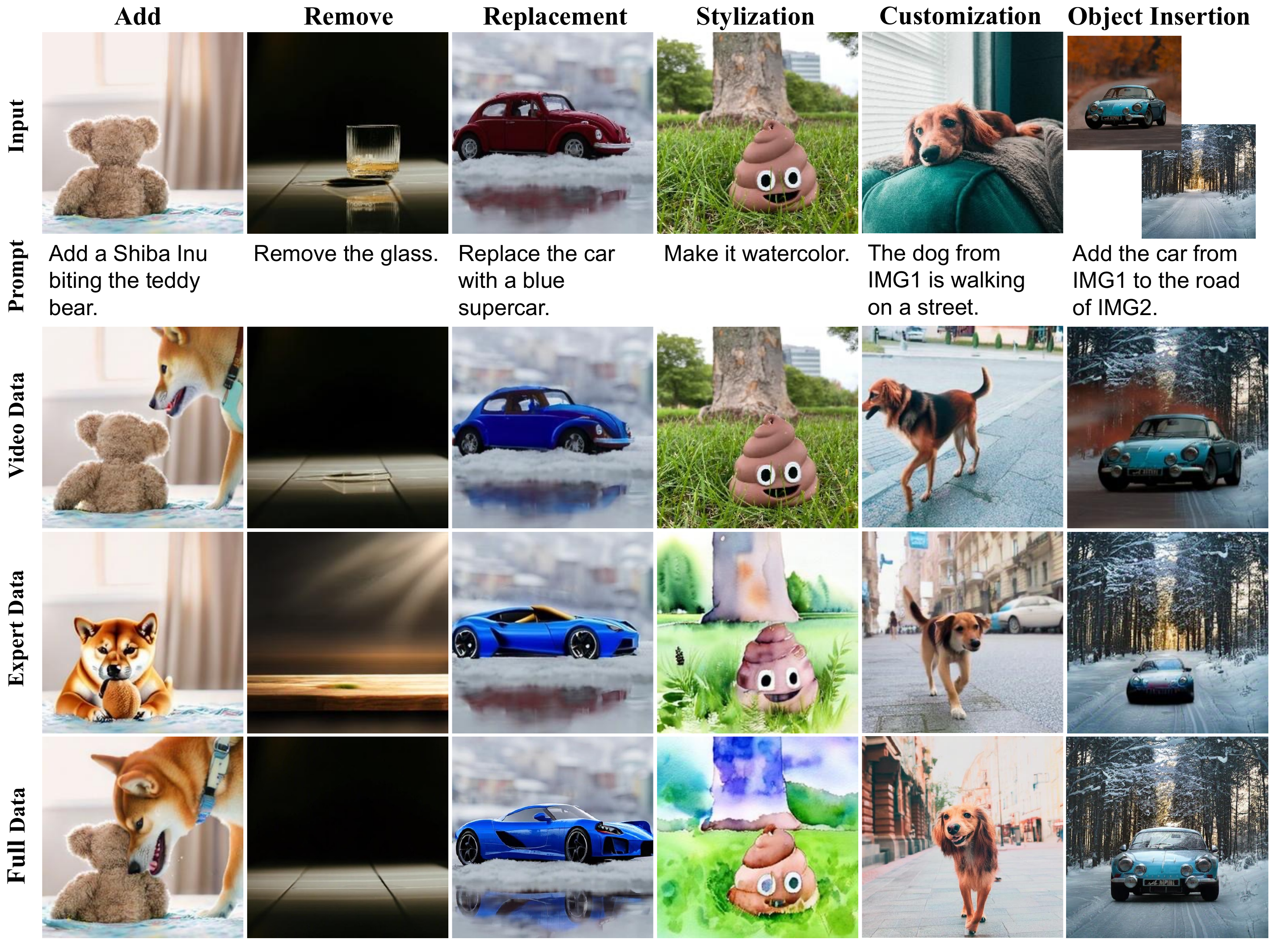} 
\vspace{-8pt}
\caption{%
    \textbf{Ablation study for the training data.} 
    We visualize the results for models that are trained on \textit{Video Frame2Frame} dataset, task-specific expert dataset, and our multi-task full dataset. 
    It is impressive that the model trained only on video data could master many editing tasks~(\textit{e.g.}, add, remove, attribute/pose changing), even for tasks with multiple input images.
}
\label{fig:ab_data}
\vspace{-5pt}
\end{figure}

\begin{figure*}[t]
\centering 
\includegraphics[width=1.0\linewidth]{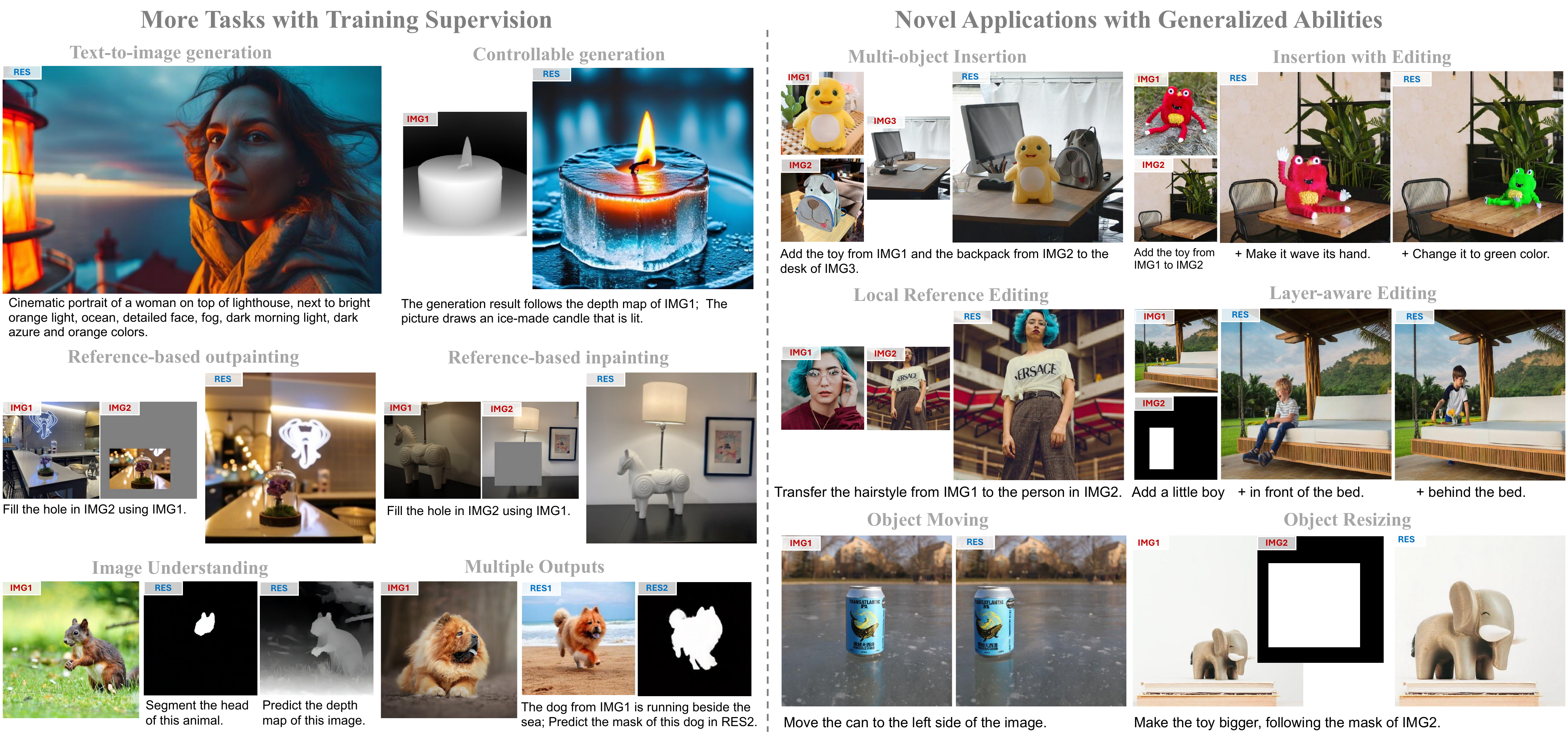} 
\vspace{-22pt}
\caption{%
    \textbf{More applications supported by \method.} 
    Our method supports a large range of applications.
    The left block demonstrates more tasks for which we constructed corresponding training data. 
    The right block shows some zero-shot novel applications achieved by task composition and generalization.
}
\label{fig:demo}
\vspace{-15pt}
\end{figure*}

\subsection{Analysis for the Core Components}
We conduct a detailed analysis of the core designs of our framework from the model and data perspectives.

\noindent \textbf{Hierachical prompt.} As introduced in \cref{sec:model}, we expand the base prompt with the image prompt and context prompt. We illustrate their effects in \cref{fig:ab_hie_prompt}, where the same input images and text prompt could lead to different types of generation targets given different image prompts~(first row) and context prompts~(second row).
With the design of hierarchical prompts, we could effectively reduce the training ambiguity when involving multiple tasks and multiple data sources. During inference, we could automatically expand the image and context prompt according to the base prompt. At the same time, users could revise the expanded prompts to better align their intentions.

\setlength{\tabcolsep}{4pt}
\begin{table}[t]
\centering
\caption{\textbf{Quantitative studies} for our basic components on MagicBrush~\cite{zhang2024magicbrush} test sets, and DreamBench~\cite{ruiz2023dreambooth}.
\vspace{-8pt}
}
\label{tab:ab_all}
\scalebox{0.65}
{
    \centering
    \begin{tabular}{lccc | ccccc}
        \toprule
         & \multicolumn{3}{c|}{ MagicBrush Test set} & \multicolumn{3}{c}{DreamBench}\\
        \midrule
        Method & $\text{CLIP}_{dir}\!\uparrow$ &   $\text{CLIP}_{out}\!\uparrow$ &  DINO$\uparrow$  & $\text{CLIP-T}\!\uparrow$ &  $\text{CLIP-I}\!\uparrow$ & $\text{DINO}\!\uparrow$ &  \\
        \midrule
         w/o Context Prompt & 0.144  & 0.294  & 0.769  & 0.315  & 0.781   &  0.683 \\
         w/o Image Prompt & 0.136  & 0.305  & 0.809   & 0.295  & 0.782   & 0.698  \\
         
         only Expert Data  & 0.139  & \textbf{0.310}  & 0.788  & 0.309  & 0.790   & \textbf{0.708}  \\
        \method-full  & \textbf{0.151} & 0.308 & \textbf{0.837} & \textbf{0.326} & \textbf{0.806}  & 0.702  \\
        \bottomrule
    \end{tabular}
}
\vspace{-15pt}
\end{table}

\noindent \textbf{Training data.} In \cref{fig:ab_data}, we compare the model only training on \textit{Video Frame2Frame}, on task-specific data~(dataset$_{1,~2,~6}$ from \cref{tab:datasets} for instructive editing, dataset$_{9}$ for customization, dataset$_{10}$ for object insertion), and on full data. 
We observe that the model trained solely on video data demonstrates a broad range of capabilities, including object add/remove, color editing, and image customization. 
Notably, although the video data contains only single input images, the trained model can handle multiple input images~(not stable), such as reference-based object insertion shown in the last column. This highlights the promising potential of video data to serve as a form of universal supervision.
Although the video data is powerful, it is still hard to cover all subtasks~(\textit{e.g.,} image stylization) and sometimes fails to precisely understand the instructions.
In this case, we still need the standard task-specific data to provide standard learning examples.

\noindent \textbf{Quantitative analysis.} The quantitative ablations are reported in \cref{tab:ab_all}. As the context prompt and image prompt could reduce the ambiguity for both training and inference, they contribute to the quantitative results. 

We also observe that the full version with multi-task training could give stronger performance than the task-specific model, which is consistent with \cref{fig:ab_data}.
We analyze that the expert data for some specific tasks may not be sufficient enough to cover all the cases or generalize well, and even could cause overfitting. 
Besides, existing instructive editing datasets leverage synthetic data, which may cause generation artifacts. In this case, the realistic video data helps the model retain generation quality. 
In general, it is not easy to collect comprehensive data, even only for a specific task. Hence, even though not aligned in target, different tasks could help each other to complement the shortage of cases, so as to improve generalization capability.

\subsection{More Applications}

In \cref{fig:demo}, the left block shows more abilities with training samples. The right part demonstrates some novel abilities.

\noindent \textbf{Trained tasks.} The first row demonstrates text-to-image and controllable generation, showcasing \method's ability to handle various aspect ratios and resolutions, producing highly aesthetic content. The second row illustrates reference-guided inpainting~\cite{tang2024realfill,zhou2021transfill,chen2024mimicbrush}, where a reference image fills the masked region of another image, preserving reference details while adapting to the target environment. The bottom row highlights \method’s capabilities in basic perception tasks like referring segmentation, depth estimation, and generating multiple output images in a single pass.

\noindent \textbf{Novel Abilities.}
\method demonstrates generalization abilities for unseen tasks. For instance, as shown in the first row, although trained on single-object insertion, it naturally supports multi-object insertion at inference. 
Additionally, it can combine tasks, such as performing object insertion with pose editing or color modification. 
The second row highlights the ability to transfer specific local features, like hairstyles,  without any input masks. 
When an editing region is specified with a mask, \method can add objects to background layers while preserving foreground elements. The final row showcases object manipulation capabilities, including moving and resizing.

\section{Conclusion}\label{sec:conclusion}
We introduce \method, a universal solution for a wide range of image generation and editing tasks.
To handle varying numbers of input and output images, we employ a video generation framework that treats images as individual frames. 
Beyond task-specific data, we leverage universal supervision from video data to learn consistency and variation across images. 
\method achieves state-of-the-art performance in multiple image generation and editing tasks and demonstrates promising capabilities in understanding real-world dynamics and generalizing to new tasks.

\noindent \textbf{Limitations.} While \method theoretically supports any number of input and output images, stability decreases, and computation becomes intensive as the number of images exceeds five. 
Typically, 3-4 input images are sufficient for most applications. 
However, to support specific tasks requiring a large number of inputs or outputs, constructing training data with more images and exploring more efficient model architectures would be necessary.
\vspace{-25pt}
{
\small
\renewcommand\UrlFont{\color{Gray}\ttfamily}
\bibliographystyle{ieeenat_fullname}
\bibliography{ref.bib}
}

\end{document}